\documentclass[journal]{IEEEtran}

\ifCLASSINFOpdf
\else
\fi

\usepackage[justification=centering]{caption} 
\usepackage{graphicx}
\usepackage{xcolor}
\usepackage{amsfonts}
\usepackage{times}

\usepackage{soul}
\usepackage{url}
\usepackage[hidelinks]{hyperref}
\usepackage[utf8]{inputenc}
\usepackage{amsmath}
\usepackage{amssymb}
\usepackage{booktabs}
\usepackage{float}
\usepackage{enumerate}
\usepackage{algorithm}
\usepackage{algorithmic}

\usepackage{multirow}
\usepackage{multicol}
\usepackage{arydshln}
\usepackage{subfigure}
\begin{document}
\colorlet{punct}{red!60!black}
\definecolor{background}{HTML}{EEEEEE}
\definecolor{delim}{RGB}{20,105,176}
\colorlet{numb}{magenta!60!black}

%

\title{Efficient Ring-topology Decentralized Federated Learning with Deep Generative Models for Industrial Artificial Intelligent}


\author{\IEEEauthorblockN{Zhao Wang\IEEEauthorrefmark{1},
Yifan Hu\IEEEauthorrefmark{2},
Jun Xiao\IEEEauthorrefmark{3}, and
Chao Wu\IEEEauthorrefmark{4}}
\IEEEauthorblockA{\IEEEauthorrefmark{1}Zhejiang University, China}
\IEEEauthorblockA{\IEEEauthorrefmark{2}Zhejiang University, China}
\IEEEauthorblockA{\IEEEauthorrefmark{3}Zhejiang University, China}
\IEEEauthorblockA{\IEEEauthorrefmark{4}Zhejiang University, China}
\thanks{Corresponding author: Zhao Wang(email: zhao\_wang@zju.edu.cn), Chao Wu (email: chao.wu@zju.edu.cn)}}

%




\IEEEtitleabstractindextext{%
\begin{abstract}
 By leveraging deep learning based technologies, the data-driven based approaches have reached great success with the rapid increase of data generated of Industrial Indernet of Things(IIot). However, security and privacy concerns are obstacles for data providers in many sensitive data-driven industrial scenarios, such as healthcare and auto-driving. Many Federated Learning(FL) approaches have been proposed with DNNs for IIoT applications, these works still suffer from low usability of data due to data incompleteness, low quality, insufficient quantity, sensitivity, etc. Therefore, we propose a ring-topogy based decentralized federated learning(RDFL) scheme for Deep Generative Models(DGMs), where DGMs is a promising solution for solving the aforementioned data usability issues. Compare with existing IIoT FL works, our RDFL schemes provides communication efficiency and maintain training performance to boost DGMs in target IIoT tasks. A novel ring FL topology as well as a map-reduce based synchronizing method are designed in the proposed RDFL to improve decentralized FL performance and bandwidth utilization. In addition, InterPlanetary File System(IPFS) is introduced to further improve communication efficiency and FL security. Extensive experiments have been taken to demonstate the superiority of RDFL with either independent and identically distributed(IID) datasets or non-independent and identically distributed(Non-IID) datasets. 
 
\end{abstract}

}

\maketitle

\IEEEdisplaynontitleabstractindextext

%
\IEEEpeerreviewmaketitle

\section{Introduction}
Recent years have witnessed a rapid growth of deep learning(DL) algorithms that widely used to solve data-driven industrial problems in real-world industrial applications. These deep learning methods are benefited a lot by the massive amount of data collected. To improve the DL based products, it brings great demand for different entities, e.g. huge amounts of IoT devices belong to different owners, to contribute their own data and train models together. In such collaborative training, the collection of massive data for centralized training causes serious privacy threats \cite{ren2018querying}, which motivated federated learning(FL) \cite{mcmahan2017communication} that allows participants to learn the model collaboratively by only synchronizing local-trained model parameters without revealing their own original data.   

A general federated learning systems usually uses central parameter server to coordinate the large federation of participating nodes(nodes, clients and workers are used interchangeably in this manuscript).  For instance, Conventional FL framework \cite{ryffel2018generic,caldas2018leaf} uses a highly centralized architecture where a centralized node collects gradients or model parameters from data nodes to update the global model. Although some FL approaches have been proposed for IIoT applications \cite{lu2019blockchain,hao2019efficient,savazzi2020federated,liu2020privacy}, the model training performance always suffer from low usability of IIoT data, such as data  incompleteness, low quality, insufficient quantity and sensitivity. Deep Generative Models(DGM) like Generative Adversarial Network(GAN) can be used to tackle the aforementioned problems. In order to meet the data privacy constraints, distributed GAN algorithms are proposed \cite{yonetani2019decentralized}. Large communication bandwidth among nodes is required in current distributed GAN algorithms\cite{heusel2017gans, augenstein2019generative}, while an intermediary is required to ensure convergence  due to its architectures which separate generators from discriminators. However, the communication bandwidth could be limited and costly in many real-world IIoT applications\cite{yang2019federated}. The centre node in current FL framework suffers  from communication pressure and communication bandwidth bottleneck \cite{tang2020communication,philippenko2020artemis}. Communication-efficient distributed GAN is still an open problem and We propose a framework that places local discriminators with local generators and synchronize occasionally. 

Additionally, the aforementioned centralized FL frameworks could bring security concerns and suffer the risk of single point failure. Through the literature review, the decentralized FL framework \cite{he2019central,li2020blockchain,hu2019decentralized} has been proposed. The decentralized FL framework removes the centralized node and synchronizes FL updates among the data nodes, then performs aggregation. However, It still faces challenges in communication pressure and cost, especially when blockchain is employed as an effective decentralized storage and replace the central FL servers \cite{zhao2019mobile,li2020blockchain}.  In addition, it's important to design the aggregation algorithm used in the decentralized FL framework that is able to achieve competitive performance under the situation of data poisoning from malicious nodes. 

To tackle aforementioned problems, a ring-topology decentralized federated learning(RDFL) framework is proposed in this paper. RDFL aims to provide communication-efficient learning across multiple data sources in a decentralize structure, which is also subject to privacy constraints. Inspired by the idea of ring-allreduce\footnote{https://andrew.gibiansky.com/blog/machine-learning/baidu-allreduce/}, Consistent hashing technique \cite{lamping2014fast} is employed in the proposed RDFL to construct a ring topology of decentralized nodes, which is able to reduce the communication pressure and improve topology stability. Besides, an innovative model synchronizing method is also designed in RDFL to benefit the bandwidth utilization and decentralized FL performance. Additionally, InterPlanetary File System(IPFS) \cite{benet2014ipfs} based data sharing scheme is also designed to further improve communication efficiency and reduce communication cost. The code of RDFL will be published online soon. 

To sum up, the main contributions of proposed RDFL are as follows:
\begin{itemize}
\item[1)] A new data node topology mechanism for decentralized FL has been designed in this work. The proposed mechanism is able to significantly reduce communication pressure and improve system stability. To the best of our knowledge, this is the first attempt to conduct data node topology design for communication-efficient decentralized FL.
\item[2)] A novel ring decentralized federated learning(RDFL) synchronizing method is designed to improve the bandwidth utilization and training stability.
\item[3)] To improve the communication performance and security of the decentralized FL framework, a IPFS based data sharing scheme is desinged to reduce system communication pressure and cost.
\end{itemize}

\section{Related Work}
This work relates to two literature, federated learning and distributed/federated GAN. 

Federated learning has emerged as a new paradigm in distributed machine learning setup \cite{mcmahan2017communication} and was widespread by Google's blog post\footnote{https://ai.googleblog.com/2017/04/federated-learning-collaborative.html}. It \cite{mcmahan2017communication} proposes a FL process that collects locally calculated gradients and aggregates them at the central node. To help build FL task, some centralized FL frameworks have been proposed. Representatives of these frameworks are FATE\footnote{https://fate.fedai.org/}, TensorFlow-Federated(TFF)\footnote{https://www.tensorflow.org/federated}, PaddleFL\footnote{https://github.com/PaddlePaddle/PaddleFL}, LEAF \cite{caldas2018leaf} and PySyft \cite{ryffel2018generic}. However, these centralized FL frameworks still have the problem of security concern, communication bottleneck and stability.

To avoid the problems caused by centralized FL framework, the research on the decentralized FL framework has attracted much attention.  In \cite{hu2019decentralized}, it has proposed a decentralized FL algorithm based on Gossip algorithm and model segmentation. Local models are propagated over a peer-to-peer network through sum-weight gossip. Roy et al. have proposed a peer-to-peer decentralized FL algorithm. Lalitha et al. have explored fully decentralized FL algorithm \cite{lalitha2018fully}. A blockchain-based decentralized FL framework is presented in \cite{li2020blockchain}. To overcome the communication problem of decentralized FL framework, current research focuses on researching novel communication compression or model compression techniques to reduce the communication pressure. For instance, Hu et al. utilizes the gossip algorithm to improve bandwidth utilization and model segmentation to reduce communication pressure\cite{hu2019decentralized} . Amiri et al. \cite{amiri2020federated} and Konečný el al. \cite{konevcny2016federated} propose model quantification methods to reduce communication pressure. Tang et al. \cite{tang2018communication} and Koloskova et al.\cite{koloskova2019decentralized} introduce communication compression methods to reduce communication pressure.  Besides, sharing datasets \cite{zhao2018federated} and knowledge distillation \cite{jeong2018communication,itahara2020distillation} are employed in FL to improve the FL performance on Non-IID dataset. To further protect the data privacy of FL, existing research focuses on several defense methods, including differential privacy \cite{geyer2017differentially} and multi-party secure computing(MPC) \cite{melis2019exploiting}. There are also reports on applying blockchain technology to decentralized FL to improve the security \cite{lu2019blockchain,kim2019blockchained,li2020blockchain}. 
Distributed GANs are proposed recently \cite{li2018multi,hardy2018gossiping}.  It propose a single generator at the intermediary and distributed discriminators in \cite{li2018multi}.  A gossip approach for distributed GAN which does not require an intermediary server is presented in\cite{hardy2018gossiping}. In order to deal with non-iid data, individual discriminator is trained separately while the centralized generator is updated to fool the weakest discriminator in \cite{yonetani2019decentralized}. All of the above works require large communications during training. Few attempts have been made to address the problem of GAN training in a FL way\cite{rasouli2020fedgan,fan2020federated,rajotte2021reducing}, while little attention has been payed on improving communication efficiency.

\section{The Proposed RDFL}
In this section, we firstly describe how the designed topology mechanism in RDFL that utilize consistent hashing algorithm to build a ring decentralized FL topology for FL nodes. Then we describe synchronizing method in RDFL. Finally, a IPFS based data sharing scheme is presented to further reduce communication cost.

\subsection{Ring Decentralized FL Topology}

\begin{figure}[h]
    \centering
     \includegraphics[width=0.4\textwidth]{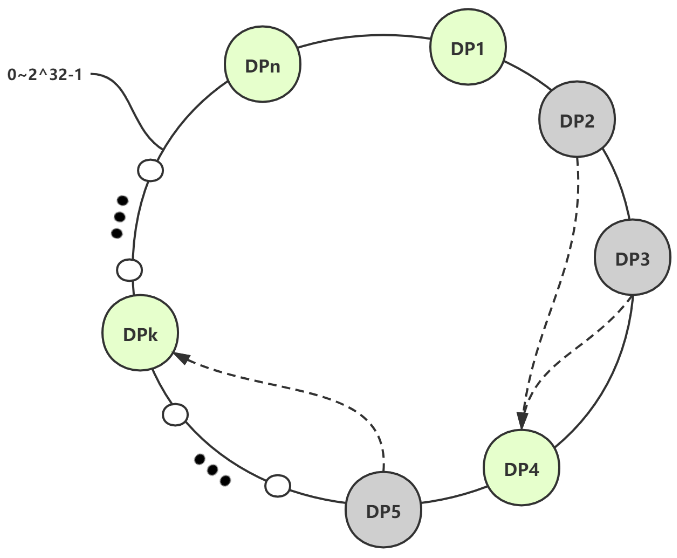}
    \caption{Ring Topology}
    \label{fig:ring_no_virtual_nodes}
\end{figure}

\textbf{Topology Overview} Consider a group of $n$ data nodes among which there are $m$ trusted data nodes and $n-m$ untrusted data nodes. These $n$ data nodes are represented by the symbol $\mathcal{D}=\{DP_1, DP_2, DP_3,...,DP_n\}$.  RDFL utilizes consistent hashing algorithm to construct a ring topology of $n$ data nodes. The consistent hash value $H_{k}=Hash(DP_{k}^{ip})\subseteq[0,2^{32}-1]$, $DP_{k}^{ip}$ represents the ip of $DP_k, k\subseteq[1,n]$. Data nodes are distributed on the ring with an index value range $[0,2^{32}-1]$ according to the consistent hash value. Figure \ref{fig:ring_no_virtual_nodes} shows the ring topology constructed by the consistent hashing algorithm.

\textbf{Malicious Node} The malicious nodes can be detected with a committee election methods\cite{li2020blockchain}. The malicious nodes will only send local models to the nearest trusted data node found with the proposed ring topology in a clockwise direction and will not be passed anymore. In Fig \ref{fig:ring_no_virtual_nodes}, the green data nodes represent trusted data nodes and the gray data nodes represent Untrusty data nodes. According to the clockwise principle, Untrusty data nodes $DP_2$ and $DP_3$ send models to trusted data provider $DP_4$. Untrusty data node $DP_5$ sends models to the nearest trusted data node $DP_k$. With the help of consistent hashing algorithm, different untrusty data nodes can only send models to its corresponding trusty nodes, which reduces the communication pressure of trusty node effectively. 

\begin{figure}[h]
    \centering
      \includegraphics[width=0.4\textwidth]{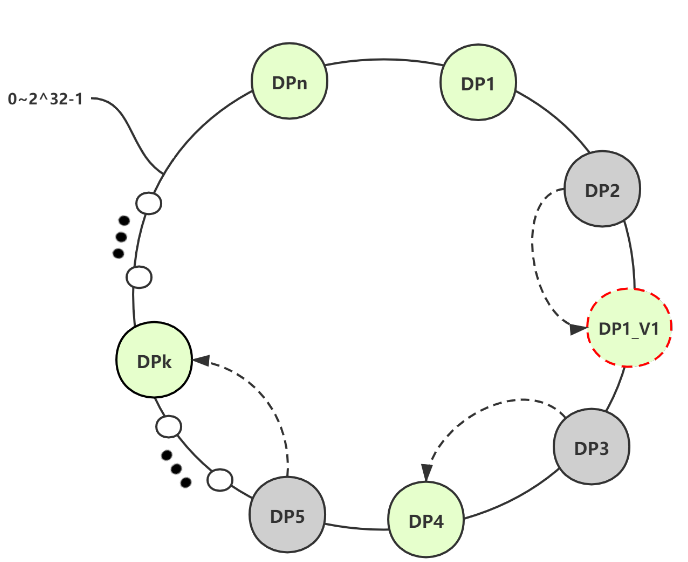}
    \caption{Ring Topology With Virtual Nodes}
    \label{fig:ring_with_virtual_nodes}
\end{figure}

In order to deal with continuous untrusty nodes, a possible solustion is making the distribution of trusty nodes on the ring uniformly. Hence, virtual nodes of trusted nodes can also be added in the ring topology if needed, which aims to further reduce communication pressure. Figure \ref{fig:ring_with_virtual_nodes} shows a ring topology with virtual nodes. The green nodes with red dashed lines represent virtual nodes. $DP_{1}^{v1}$ is the virtual node of $DP_1$. 

\subsection{RDFL training}
\begin{figure}[h]
    \centering
    \includegraphics[width=0.45\textwidth]{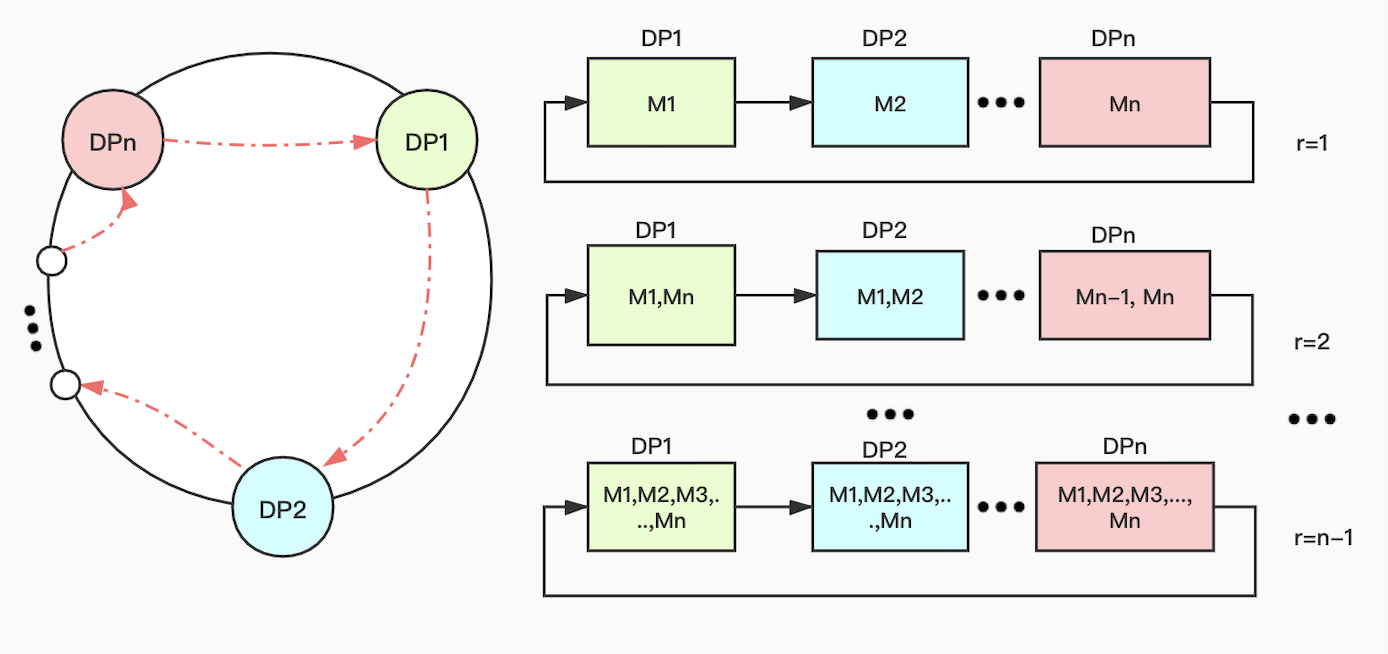}
    \caption{Ring Decentralized Federated Learning}
    \label{fig:rdfl}
\end{figure}
\textbf{Synchronizing progress} Based on the ring decentralized topology constructed with the consistent hash algorithm, the trusty node follows the synchronizing progress that illustrated in Figure \ref{fig:rdfl}. $M_{i}$ represents the model of data node $DP_{i}$. $r$ represents the number of rounds to execute models synchronization and $m$ represents the number of trusted nodes. At each synchronizing round, each node send its models in a clockwise direction, then execute \textit{Federated Averaging}(FedAvg) \cite{mcmahan2017communication} to generate a new global model and starts next iteration. 

\textbf{Training progress with GAN Models} 
The horizon training iteration is denoted by $T$ and index time is denoted by $t$. Consider nodes $DP_i, i \in {1,2,\cdots,N}$ with local dataset $R_i$ for each node, the weight of node $DP_i$ is denoted by $p_i$. Assume each node has local discriminator and generator with corresponding parameters $d^i$ and $g^i$, loss function $L^i_D$ and $L^i_G$, local stochastic gradients $\tilde{\theta}^i(d^i_t,g^i_t)$ and $\tilde{h}^i(d^i_t,g^i_t)$ and learning rate $lr^d(t)$ and $lr^g(t)$ at time $t$. We assume the learning
rates are the same across nodes. To improve the bandwidth utilization between trusted nodes, RDFL introduces the Ring-allreduce algorithm and the clockwise principle.  As show in Figure \ref{fig:rdfl}, the trusted node $DP_{1}$ retains the local model $M_1$ after distillation, $DP_{2}$ retains $M_2$ and $DP_{n}$ retains $M_n$. Then, the trusted nodes utilize the Ring-allreduce algorithm and the clockwise rule to synchronize the local models of the trusted nodes. After synchronization, all trusted nodes have local models of other trusted nodes. The detail of RDFL training is described in Algorithm \ref{alg:algorithm}.

\begin{algorithm}[tb]
\caption{RDFL training with Generative Adversarial Network}
\label{alg:algorithm}
\textbf{Input}: Set training period $T$, synchronizing interval $K$. Initialize global discriminator and generator $d_{0}$ and $g_{0}$. Initialize local discriminator and generator parameters $d_0^i = d_0$, $g_0^i = g_0$ for all $N$ nodes $DP_i, \forall i \in \{ 1,2,\cdots,N\}$.\\
\textbf{Output}: The New global model \\
\textbf{Procedure}: Data Node Executes \\
\begin{algorithmic}[1] 
\FOR{each FL round $t=1, 2, 3, . . . , T$}
\STATE Each node calculates local stochastic gradient $ \tilde{\theta}^i_t $ and $ \tilde{h}^i_t $ corresponding to local discriminator and local generator respectively while fake generated by the local generator. 
\STATE Each updates its local parameter in parallel;
$$ \left\{
\begin{array}{rcl}
    d_t^i \leftarrow d_{t-1}^i + {lr^d(t)} \tilde{\theta}^i_t \\
    g_t^i \leftarrow g_{t-1}^i + {lr^g(t)} \tilde{h}^i_t
\end{array}\right. $$

\IF{{$t \mod K = 0$}}
\STATE Malicious node detection 
\STATE  Each trusted node receive all trusty nodes' model parameters through the ring;
\STATE $B$ is the subset of $N$ stands for trusty nodes 
\STATE for $DP_{i \in B}$ executes global model parameters
    $$
    \begin{array}{rcl}
    d_t \triangleq \sum_{j=1}^B p_j d_t^j \\
    g_t \triangleq \sum_{j=1}^B p_j g_t^j
    \end{array}
    $$
\STATE Each node updates its local model parameters with the executed global parameters;
\ENDIF
\ENDFOR
\end{algorithmic}
\end{algorithm}

\begin{table*}[]
    \centering
    \begin{tabular}{c|c|c|c}
     decentralized FL Framework   & Communication Times/round & Node Pressure (MB/c) & Total Transferred Data Volume per round (MB)\\
     P2P  & 1 & $N \times M$ & $N^2M$ \\
     FL Gossip\cite{hu2019decentralized} & $round(\frac{N-1}{2})$ & $2M$ & $2NM\times round(\frac{N-1}{2})$\\
     \textbf{RDFL} & $N-1$ & $M$ & $N(N-1)M$
    \end{tabular}
    \caption{Communication Complexity Analysis}
    \label{table:communication_complexity_analysis}
\end{table*}

It needs to be pointed out that we assume all nodes on the ring participate in the communication
process. If part of the nodes meet communication failures during parameters sending, an extension work could be taken by following \cite{konevcny2016federated}. Due to the paragraph limitation, the proof of convergence for model averaging would not be listed here, a similar proof could be referred in \cite{rasouli2020fedgan}, which conduct a centralized FL framework with GAN.

\subsection{IPFS based data sharing scheme}
In the most decentralized FL work \cite{he2019central,li2020blockchain,hu2019decentralized,lu2019blockchain,kim2019blockchained,rasouli2020fedgan}, we noticed that the model parameters are transferred among data nodes directly, which occupies a lot of communication overhead and could causes serious communication cost when blockchain is employed. For instance, \textit{gas fee} is required in popular blockchain \textit{Ethereum}, where the cost could be significant high for large models. In order to reduce the risk of communication cost, an IPFS based data sharing scheme is designed in RDFL. Data files,e.g. model parameters, in IPFS would be divided into multiple pieces stored on different nodes and IPFS will generate the IPFS hash corresponding to the file. The IPFS hash is a 46 bytes string and the corresponding file can be obtained from IPFS through the IPFS hash.
\begin{figure}[h]
    \centering
    \includegraphics[width=0.49\textwidth]{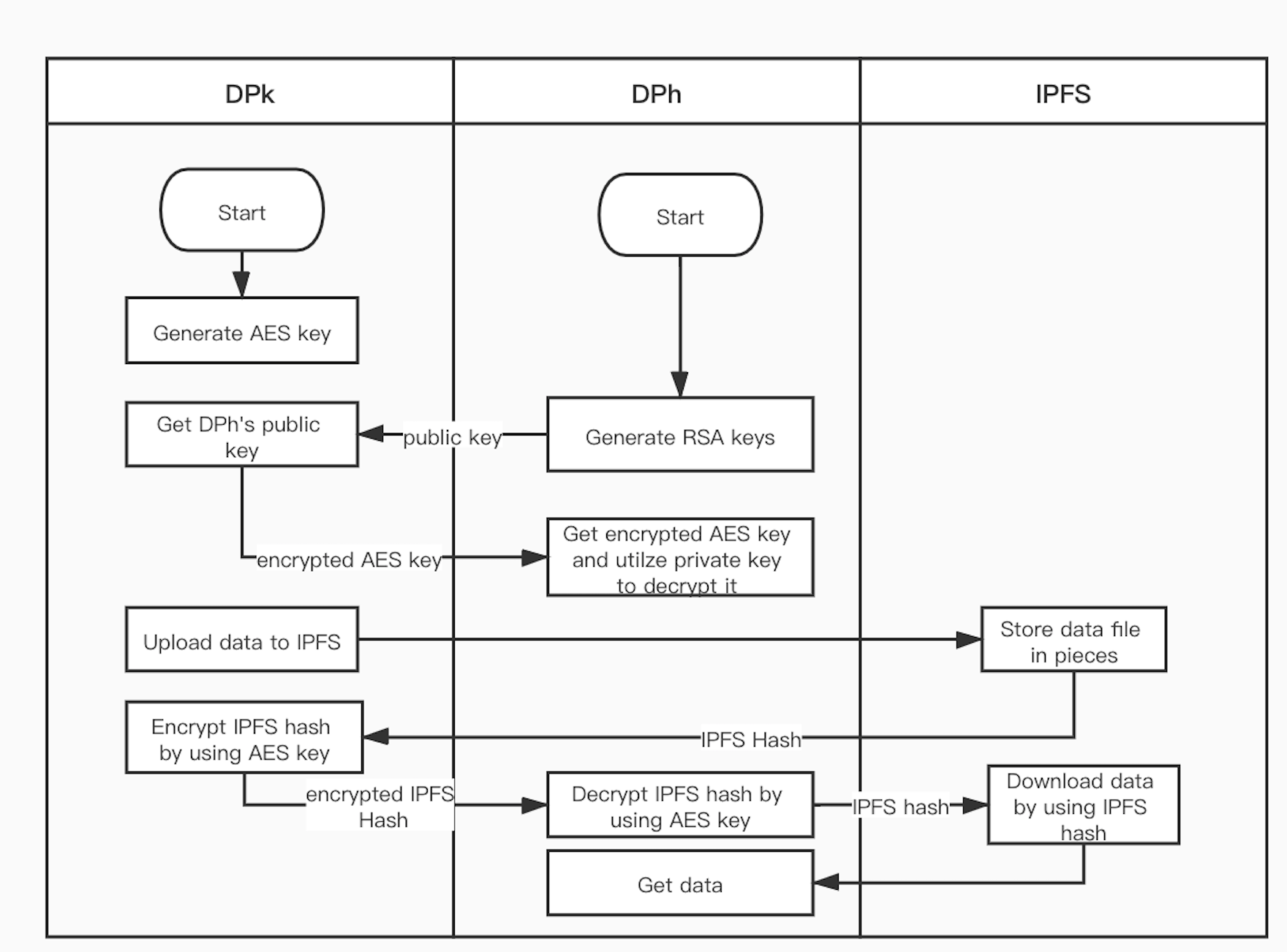}
    \caption{IPFS data sharing scheme}
    \label{fig:ipfs}
\end{figure}
As shown in Figure \ref{fig:ipfs}, data provider (node $DP_k$) send its data, e.g. model parameters, to data receiver (node $DP_h$):
\begin{enumerate}
    \item Data provider creates an AES key.
    \item Data provider stores data onto the IPFS and get the corresponding IPFS hash.
    \item Data provider encrypts the AES key in the above step using the public RSA key provided by data receiver, which ensures that only the data receiver can conduct decryption to access the AES key.
    \item \textbf{Data provider send encrypt AES key to data receiver}.
    \item \textbf{Data provider send encrypt IPFS hash to data receiver}.
    \item Data receiver get the encrypt AES key and conduct decryption with its RSA private key
    \item Data receiver get the encrypt IPFS hash and conduct decryption with received AES key that getting in the above step
    \item Data receiver get the relevant file from IPFS with the IPFS hash. 
\end{enumerate}
The direct communication between data node $DP_k$ and $DP_h$ only occurs at step 4 and 5 in the proposed scheme, where the size of both AES key and IPFS hash are significant smaller than DGM or DNN model parameters. Therefore, the proposed IPFS data sharing scheme in RDFL is able to significantly benefit the system communication efficient and reduce communication cost especially when blockchain technique is used.    

\subsection{Communication and Computation Complexity}
\begin{figure}
    \centering
    \includegraphics[width=0.24\textwidth]{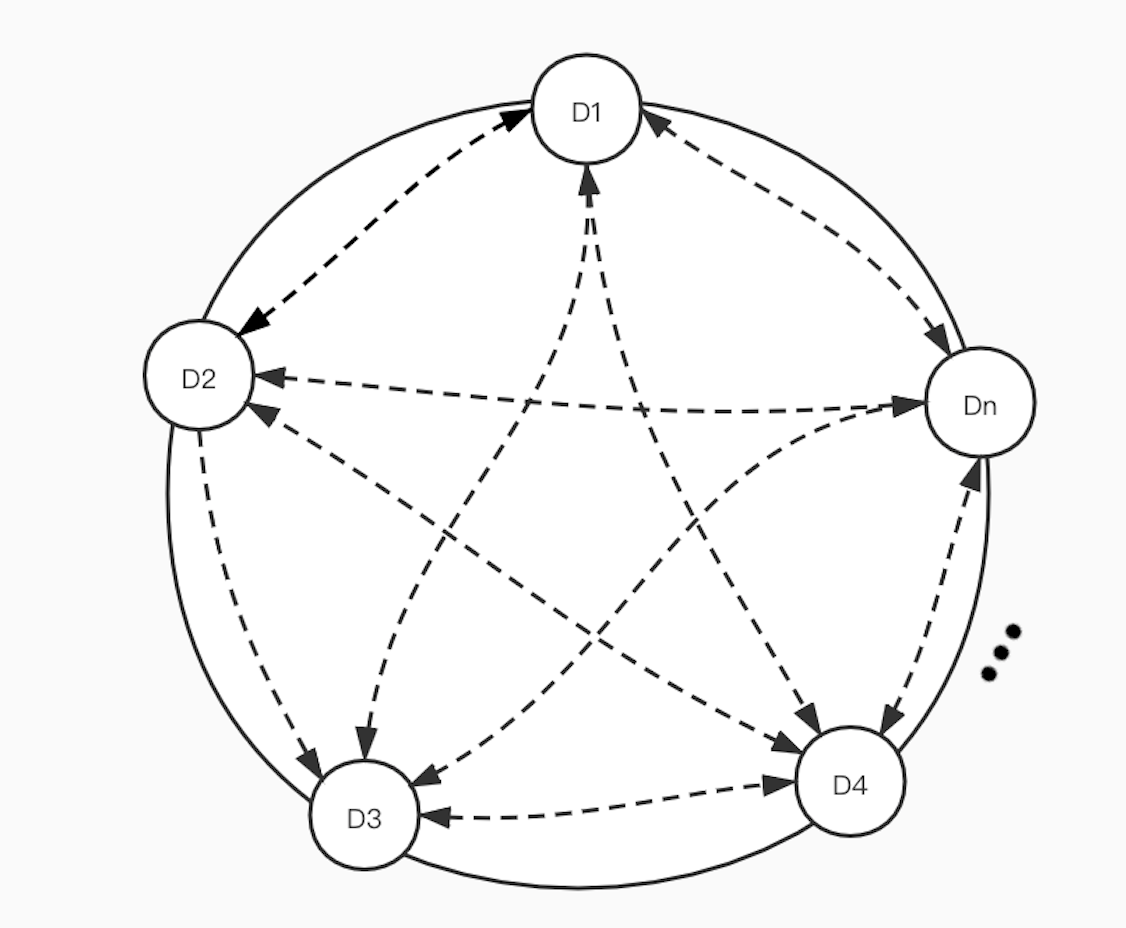}
    \includegraphics[width=0.24\textwidth]{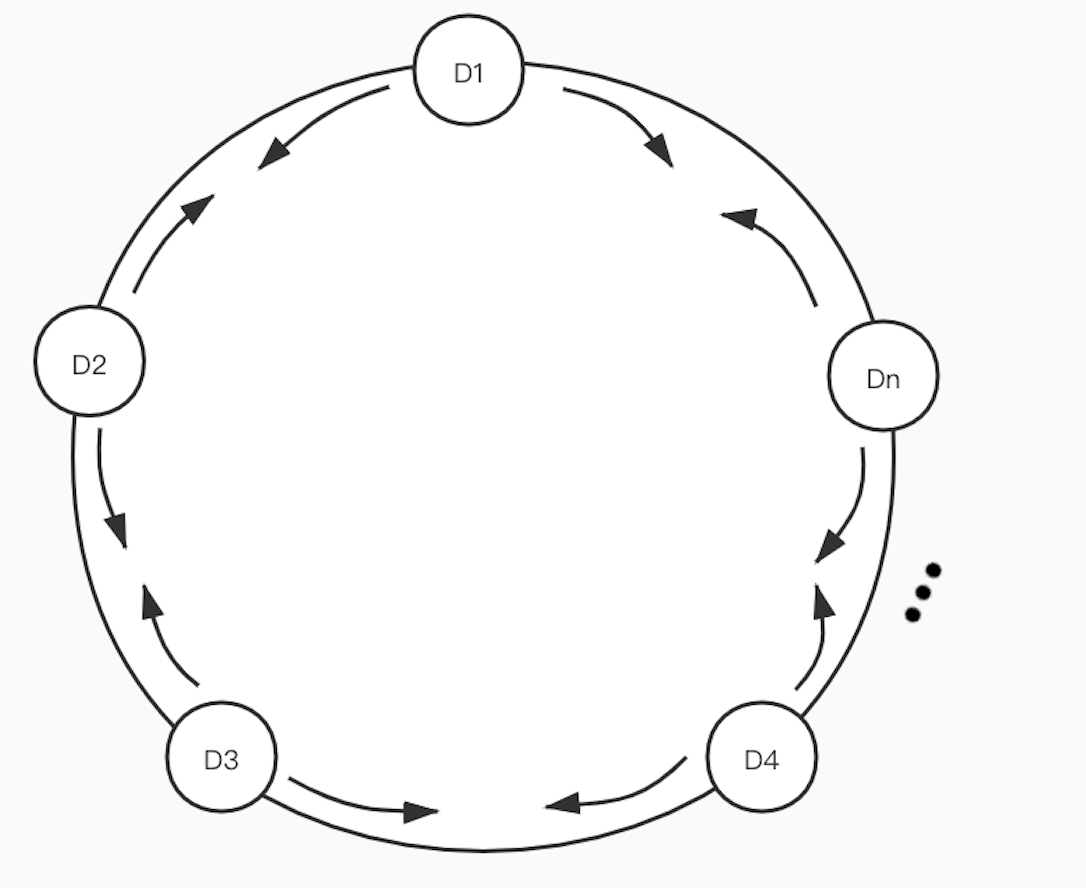}
    \caption{Overview of Decentralized FL communication methods: p2p (left) and FL Gossip(right)}
    \label{fig:overview_dfl}
\end{figure}
Since each node needs to trains its local discriminator and generator, RDFL requires similar computations compared to FedGan \cite{rasouli2020fedgan} and increased computations (roughly doubled) for each node compared to distributed GAN \cite{hardy2018gossiping}.The communications in the proposed RDFL are manily limited to parameters transferring among all nodes in each round for every $K$ steps.  Assume $M$ is the size of model parameters including discriminator and generator, there would be $N-1$ times communication in one round and the average load per communication time per node is $M$. Increasing $K$ could reduce the communication frequency, which may reduce the performance of FL train. An overview of another two decentralized FL communication methods are shown in Figure \ref{fig:overview_dfl}. A summary of communication complexity comparison is shown in table \ref{table:communication_complexity_analysis}. Generally, the total transferred data volume per FL round is similar for all three methods, The proposed RDFL achieve a better performance on communication pressure of nodes, which could benefit the system's bandwidth utilization and increase robustness.     

\section{Experimental Results and Discussions}

\begin{figure*}
    \centering
    \includegraphics[width=0.24\textwidth]{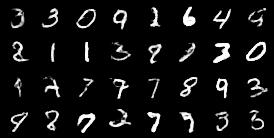}
    \includegraphics[width=0.75\textwidth]{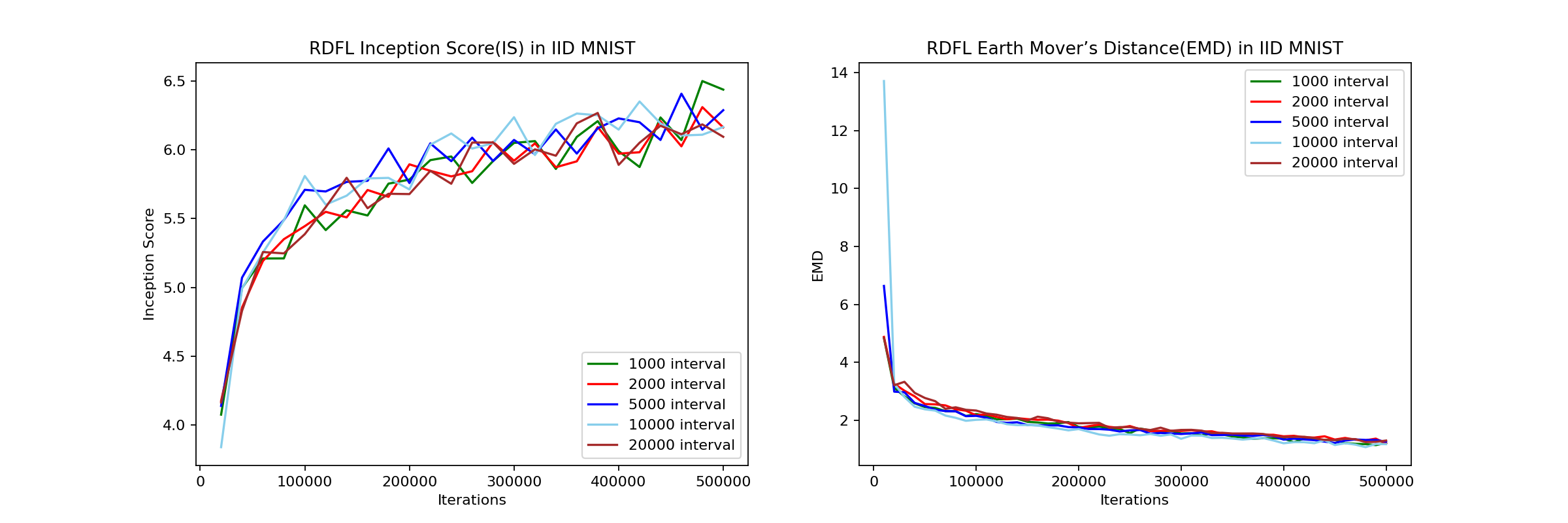}
    \caption{Illustration of FL training quality with GAN on MNIST IID, where number of nodes $B = 5$. (left) Generated images on $K=2000$, (middle) IS vs.Iterations with $K \in [1000,2000,5000,10000,20000]$, (right) EMD vs.Iterations with $K \in [1000,2000,5000,10000,20000]$}
    \label{fig:MNIST_iid}
\end{figure*}

\begin{figure*}
    \centering
    \includegraphics[width=0.24\textwidth]{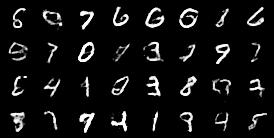}
    \includegraphics[width=0.75\textwidth]{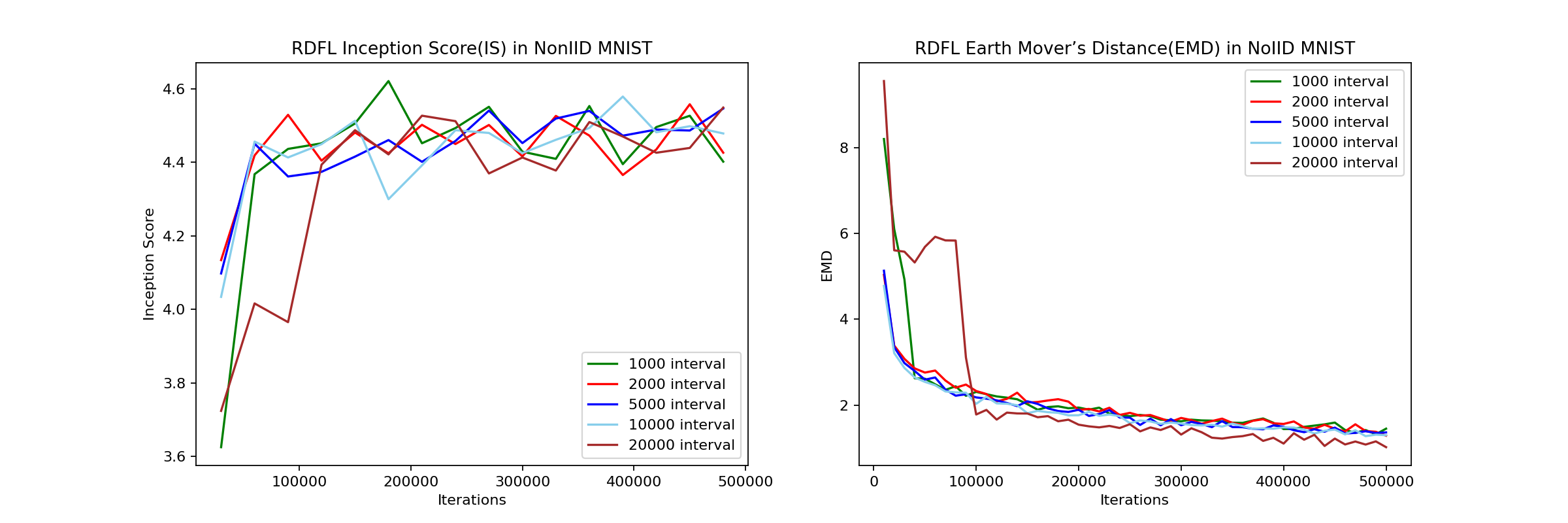}
    \caption{Illustration of FL training quality with GAN on MNIST non-IID, where number of nodes $B = 5$. (left) Generated images on $K=2000$, (middle) IS vs.Iterations with $K \in [1000,2000,5000,10000,20000]$, (right) EMD vs.Iterations with $K \in [1000,2000,5000,10000,20000]$}
    \label{fig:MNIST_noniid}
\end{figure*}

In this section, we conduct several experiment to evaluate the proposed RDFL to show its convergence, performance in generating close to real data, and robustness to reducing communications (by increasing synchronization interval $K$).
Inception Score(IS) is used in this paper, which is a common criteria in measuring the performance of GAN \cite{heusel2017gans}. Another criteria used is Earth Mover’s Distance (EMD),which is also known as Wasserstein distance. In practice, EMD is approximated by comparing average softmax scores of drawn samples from real data against the generated data such that
\begin{equation}
    EMD((x_r,y_r),(x_g,y_g)) = \frac{1}{N}\sum_{i=1}^N(f_o(x_r^i)|y_r^i|-(f_o(x_g^i)|y_g^i|)
    \nonumber
\end{equation}
where $(x_r,y_r)$ are real data samples, $(x_g,y_g)$ are generated data samples, $f_o$ is the oracle classifier mentioned above. EMD measures a relative distance between real data and fake data. Obviously, a better generator should have a lower EMD by producing realistic images closer to real images.

We build the training set of each client by randomly choosing 50
the total training samples with replacement to simulate IID data. In order to further examine the performance of RDFL on non-iid dataset, the Latent Dirichlet Allocation(LDA) and label partition method is applied to divide dataset into $N$ partitions \cite{he2020fedml}. 

\subsection{RDFL training performance with GAN}
We test RDFL on MNIST to show its performance on image datasets. MNIST  consist of 10 classes of data which we split across B = 5. The hyperparameters of GAN model we used in this experiment is listed in table \ref{table:gan_hyperparameters}. 
\begin{table}[]
\begin{tabular}{clllccccc}
\multicolumn{4}{l}{Operation}              & \multicolumn{1}{l}{Kernel} & \multicolumn{1}{l}{Strides} & \multicolumn{1}{l}{Feature maps} & \multicolumn{1}{l}{BN?} & \multicolumn{1}{l}{Non-linearity} \\
\multicolumn{9}{l}{$G(z) 100\times 1\times 1$ input}                                                                                                                                                         \\
\multicolumn{4}{c}{Trans Conv} & 4 x 4                      & 1 x 1                       & 256                              & Y                       & ReLU                             \\
\multicolumn{4}{c}{Trans Conv} & 4 x 4                      & 2 x 2                       & 128                              & Y                       & ReLU                             \\
\multicolumn{4}{c}{Trans Conv} & 4 x 4                      & 2 x 2                       & 64                               & Y                       & ReLU                             \\
\multicolumn{4}{c}{Trans Conv} & 4 x 4                      & 2 x 2                       & 3                                & N                       & Tanh                             \\
\multicolumn{9}{l}{D(x) - 32 x 32 x 3 input}                                                                                                                                                          \\
\multicolumn{4}{c}{Conv}            & 4 x 4                      & 2 x 2                       & 32                               & Y                       & LeakyReLU                        \\
\multicolumn{4}{c}{Conv}            & 4 x 4                      & 2 x 2                       & 64                               & Y                       & LeakyReLU                        \\
\multicolumn{4}{c}{Conv}            & 4 x 4                      & 2 x 2                       & 128                              & Y                       & LeakyReLU                        \\
\multicolumn{4}{c}{Conv}            & 4 x 4                      & 1 x 1                       & 1                                & N                       & -                               
\end{tabular}
\caption{GAN hyperparameters. The learning rates for generator and discriminator are both equal to the same value, across all cases.BN stands for batch normalization, Trans Conv stands for transpose convolution, Conv stands for convolution}
\label{table:gan_hyperparameters}
\end{table}

From Figure \ref{fig:MNIST_iid}, the trained GAN with RDFL is able to generate close to real images. We check Gan with RDFL performance robustness to reduced communications and increased synchronization intervals $K$ by setting $K =
1000; 2000; 5000; 10000; 20000$. The results shown in the middle and right part of Figure \ref{fig:MNIST_iid} This indicates that Gan with RDFL has high performance for image data, and furthermore its performance is robust to reducing the communications by increasing synchronization intervals $K$.

\begin{table*}
\centering

\caption{Comparison of accuracy between RDFL and FedAvg, nodes number $B=5$}
\begin{tabular}{|c|c|cc|cc|}
\hline
\multirow{2}{*}{Experiments} &
  \multirow{2}{*}{\begin{tabular}[c]{@{}c@{}}Data Node Allocation\\ (Trusty : Malicious)\end{tabular}} &
  \multicolumn{2}{c|}{FedAvg} &
  \multicolumn{2}{c|}{RDFL(ours)} \\ \cline{3-6} 
 &
   &
  \multicolumn{1}{l}{CIFAR-10} &
  \multicolumn{1}{l|}{CIFAR-100} &
  \multicolumn{1}{l}{CIFAR-10} &
  \multicolumn{1}{l|}{CIFAR-100} \\ \hline
\multirow{4}{*}{Data Poisoning(IID)} & 2:3 & 79.19 & 43.27 & \textbf{84.25} & \textbf{53.86} \\
                                     & 3:2 & 80.68 & 50.43 & \textbf{84.86} & \textbf{54.62} \\
                                     & 4:1 & 82.32 & 51.48 & \textbf{84.71} & \textbf{54.53} \\
                                     & 5:0 & 82.76 & 52.38 & \textbf{85.14} & \textbf{55.28} \\ \hline
Non-IID(LDA)                         & 4:1 & 61.72 & 32.29 & \textbf{63.25}          & \textbf{33.85}          \\ \hline
\end{tabular}
\label{table:rdfl_fedavg}
\end{table*}

Next, we conduct the experiment for GAN with RDFL under non-iid scenario. The results is shown in Figure \ref{fig:MNIST_noniid}. It could be seen that the GAN with RDFL could still finish training with acceptable image generation quality. We would like to encourage researchers to tackle the problem of federated learning of GANs with non-IID data in
future.

\subsection{Malicious Node Defencing}
In order to explore the robustness of RDFL again with malicious node quantitatively. We further design a classification experiment with cifar-10 and cifa-100 datasets. 

\section{Conclusion}
In this paper, we propose a decentralized FL framework based for DGMs called RDFL to tackle the problems in existing decentralized FL frameworks. RDFL utilizes consistent hashing algorithm and Ring-allreduce to improve communication performance, decentralized FL performance and stability. Moreover, RDFL introduces IPFS further improve communication performance and reduce communication cost. We hope that RDFL can help more decentralized FL with DGMs on IIoT applications and the code will be published online soon. The experimental result is shown in Table \ref{table:rdfl_fedavg}. We examine the classification performance with different node allocation. The ratio for number of trusty nodes vs. the number of malicious nodes is varied in the range of $[2:3;3:2;4:1;5:0]$ under iid scenario and $[4:1]$ under non-iid scenario. The result shows that the proposed RDFL outperform the normal FedAvg in all cases, which indicate the effectiveness of the designed ring-toplogy and synchronizing mechanism.   

\ifCLASSOPTIONcaptionsoff
  \newpage
\fi



%



\bibliography{references}{}
\bibliographystyle{IEEEtran}

%









\end{document}